\documentclass[letterpaper,10pt,journal,twoside]{IEEEtran}
\usepackage{amsmath,amsfonts}
\usepackage{algorithmic}
\usepackage{algorithm}
\usepackage{array}
\usepackage[caption=false,font=normalsize,labelfont=sf,textfont=sf]{subfig}
\usepackage{textcomp}
\usepackage{stfloats}
\usepackage{url}
\usepackage{verbatim}
\usepackage{graphicx}
\usepackage{cite}
\usepackage{cleveref}
\usepackage{booktabs}
\usepackage{multirow}
\usepackage{bbm}
\usepackage{xcolor}

\begin{document}

\title{VFM-Recon: Unlocking Cross-Domain Scene-Level Neural Reconstruction with Scale-Aligned Foundation Priors}

\author{
Yuhang Ming$^{1,2,*}$,~\IEEEmembership{Member,~IEEE,}
Tingkang Xi$^{1,*}$,
Xingrui Yang$^{3,*}$,
Lixin Yang$^{2}$,
Yong Peng$^{1,\dagger}$,~\IEEEmembership{Senior Member,~IEEE,}
Cewu Lu$^{2}$,~\IEEEmembership{Senior Member,~IEEE,}
and Wanzeng Kong$^{1}$,~\IEEEmembership{Senior Member,~IEEE}%
\thanks{This work was supported in part by the National Natural Science Foundation of China under Grant 62401188, the Zhejiang Provincial Natural Science Foundation of China under Grant LQN25F030015, the Central Government-Guided Local Science and Technology Development Fund under Grant 2025ZY01045 and 2026ZY01008, and the Zhejiang Provincial Key Laboratory of Brain Computer Collaborative Intelligence Technology and Applications under Grant 2025E10015. 
}%
\thanks{$^{1}$Yuhang Ming, Tingkang Xi, Yong Peng, and Wanzeng Kong are with the School of Computer Science, Hangzhou Dianzi University, and the Zhejiang Provincial Key Laboratory of Brain Computer Collaborative Intelligence Technology and Applications, Hangzhou, China. (email: \{yuhang.ming, tingkangxi, yongpeng, kongwanzeng\}@hdu.edu.cn.)}%
\thanks{$^{2}$Yuhang Ming, Lixin Yang, and Cewu Lu are with the School of Artificial Intelligence, Shanghai Jiao Tong University, Shanghai, China. (email: \{yohann, siriusyang, lucewu\}@sjtu.edu.cn)}%
\thanks{$^{3}$Xingrui Yang is with CARDC, Mianyang, China. (email: yangxingrui@cardc.cn)}%
\thanks{$^{*}$ denotes equal contribution. $^{\dagger}$ denotes corresponding author.}%
\thanks{The project page is available at https://whole2.github.io/VFM-Recon-Project-Page/}
}
\IEEEaftertitletext{\vspace{-\baselineskip}}



\maketitle

\begin{abstract}
Scene-level neural volumetric reconstruction from monocular videos remains challenging, especially under severe domain shifts.
Although recent advances in vision foundation models (VFMs) provide transferable generalized priors learned from large-scale data, their scale-ambiguous predictions are incompatible with the scale consistency required by volumetric fusion. 
To address this gap, we present \textbf{VFM-Recon}, \textit{the first attempt} to bridge transferable VFM priors with scale-consistent requirements in scene-level neural reconstruction.
Specifically, we first introduce a lightweight scale alignment stage that restores multi-view scale coherence. 
We then integrate pretrained VFM features into the neural volumetric reconstruction pipeline via lightweight task-specific adapters, which are trained for reconstruction while preserving the cross-domain robustness of pretrained representations.
We train our model on ScanNet train split and evaluate on both in-distribution ScanNet test split and out-of-distribution TUM RGBD and Tanks and Temples datasets. The results demonstrate that our model achieves state-of-the-art performance across all dataset domains. 
In particular, on the challenging outdoor Tanks and Temples dataset, our model achieves an F1 score of \textit{73.5} in reconstructed mesh evaluation, substantially outperforming the closest neural volumetric reconstruction competitor, SimpleRecon, with an F1 score of \textit{35.7}, as well as the closest VFM competitor, $\pi^3$, with an F1 score of \textit{67.0}.
\end{abstract}


\section{Introduction}
\label{sec:intro}

3D volumetric reconstruction from monocular RGB video aims to construct a unified geometric representation of an entire scene that is globally consistent across all views~\cite{zuo2023ral, georecon2025ral}.
It is a fundamental capability to support robotic applications like navigation and interactions. Classical approaches build such representations through a depth-estimation-and-fusion paradigm~\cite{kinectfusion2011ismar, fdslam2022icra}. While theoretically grounded, these geometry-driven methods depend heavily on sufficient view overlap and robust feature matching.

\begin{figure*}[t]
    \centering
    \includegraphics[width=\linewidth]{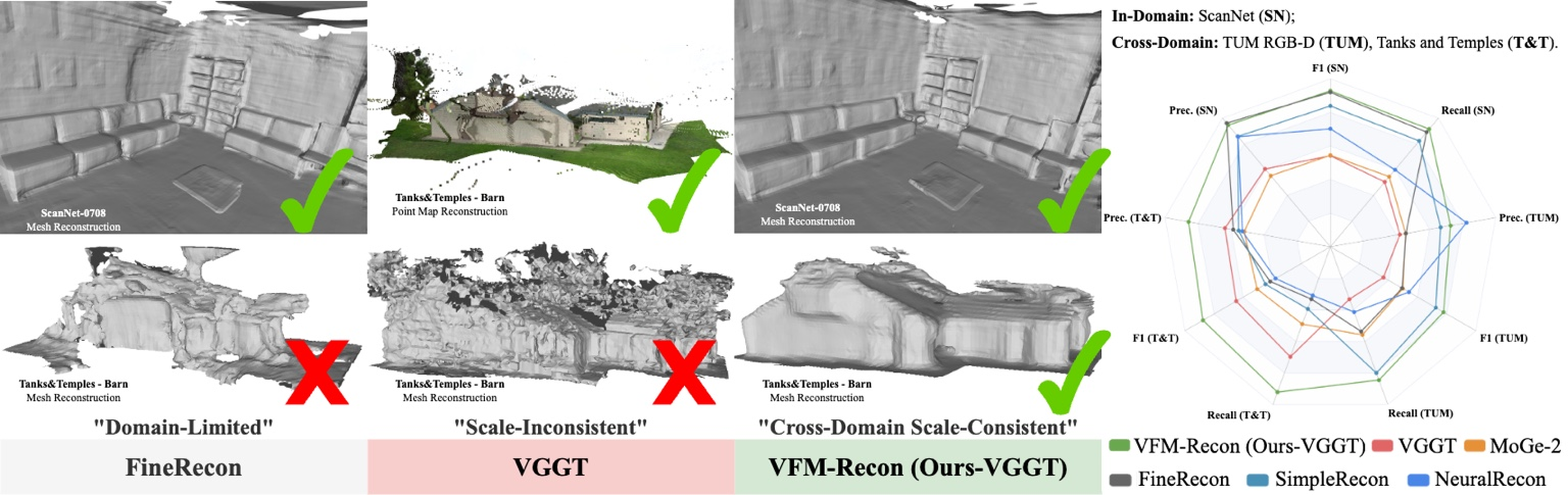}
    \caption{We present \textbf{VFM-Recon}, a novel framework that enables cross-domain, scale-consistent neural volumetric reconstruction. Existing neural reconstruction methods such as FineRecon~\cite{finerecon2023iccv} perform well on indoor scenes but struggle under domain shift, while VFMs, taking VGGT~\cite{vggt2025cvpr} for example, predict plausible point maps, yet lack scale consistency for mesh reconstruction. The radar plot highlights consistent performance gains across ScanNet~\cite{scannet2017cvpr}, TUM RGBD~\cite{tumrgbd2012iros}, and Tanks and Temples~\cite{tat2017tog}.}
    \label{fig:teaser}
\end{figure*}

To overcome these limitations, recent learning-based approaches replace explicit multi-view stereo with learned feature aggregation and volumetric decoding, enabling temporally coherent reconstruction from RGB video~\cite{neuralrecon2021cvpr, finerecon2023iccv, cdrnet2024tcsvt}. By implicitly aggregating multi-view cues within convolutional encoders and volumetric decoders, these methods achieve strong performance on indoor benchmarks~\cite{scannet2017cvpr}. However, unlike classical approaches that explicitly enforce geometric constraints, neural volumetric reconstruction systems primarily rely on learned scene priors~\cite{finerecon2023iccv}. Their performance is therefore tightly coupled to the training distribution and degrades under domain shifts, such as transferring from indoor to outdoor environments, as shown in \Cref{fig:teaser}.

In parallel, vision foundation models (VFMs), pretrained on large-scale data, demonstrate robust cross-domain generalization across diverse tasks~\cite{dav22024neurips, dino2024tmlr, vggt2025cvpr, pi32026iclr}. Recent studies reveal that these models encode powerful appearance and geometric priors~\cite{eval3d2025unireps}, which can be used to enhance downstream tasks~\cite{gsloc, dcvlaq2026arxiv, colonadapter}. 
We believe geometry-aware VFMs like VGGT~\cite{vggt2025cvpr} and $\pi^3$~\cite{pi32026iclr} provide two complementary signals for neural volumetric reconstruction: explicit geometric predictions such as monocular depth, and transferable geometry-aware feature embeddings that can strengthen multi-view aggregation. It is worth noting that unlike recent large reconstruction models~\cite{lrm2024iclr, sf3d2025cvpr} that focus on object-level reconstruction, we are, to the best of our knowledge, the first to systematically leverage VFM priors to enhance scene-level neural volumetric reconstruction under domain shifts.

Yet directly injecting VFM priors into neural volumetric reconstruction is non-trivial. Although VFM-predicted depth captures strong relative structure, it remains scale-ambiguous and may exhibit inter-frame inconsistency over long sequences~\cite{vggtslam2025neurips}. In contrast, volumetric fusion requires scale consistent multi-view observations~\cite{finerecon2023iccv, dgrecon2023iccv}; even small scale mismatches accumulate during fusion, leading to unstable geometry, as shown in \Cref{fig:teaser}. This exposes a structural incompatibility: VFMs provide transferable geometric priors without scale guarantees, whereas volumetric reconstruction critically depends on scale-consistent integration.

In this work, we propose a novel framework that reconciles VFM priors with scale-consistent neural volumetric reconstruction through a decoupled design. We introduce \textit{a lightweight scale alignment stage} that resolves global and local scale ambiguity in VFM depth using relative camera poses and pairwise geometric consistency. Rather than re-optimizing trajectories or dense volumes, we estimate a compact set of scale factors to restore multi-view coherence, enabling stable volumetric fusion. Then, in the \textit{VFM-augmented volumetric reconstruction stage}, we integrate task-adapted VFM feature embeddings into the volumetric reconstruction network. These features act as geometry-aware structural priors, improving cross-domain robustness while preserving the scale stability required for consistent scene-level fusion.

To validate our design, we conduct extensive experiments on \textit{ScanNet}~\cite{scannet2017cvpr} using the official train/test splits. To evaluate cross-domain generalization, we further test on out-of-distribution scenes from \textit{TUM RGBD}~\cite{tumrgbd2012iros} and \textit{Tanks and Temples}~\cite{tat2017tog}. Our model achieves state-of-the-art (SoTA) performance across all domains. 

In summary, our contributions are threefold:
\begin{enumerate}
    \item 
    We are, to the best of our knowledge, \textbf{the first} to systematically leverage VFM priors to enhance \textbf{scene-level neural volumetric reconstruction} under domain shifts, bridging transferable foundation representations with scale-consistent fusion requirements.

    \item We propose a \textbf{lightweight scale alignment module} that resolves global and local scale ambiguity using geometric consistency without full scene re-optimization.
    
    \item We introduce a \textbf{VFM-augmented volumetric reconstruction network} that significantly improves cross-domain generalization while maintaining reconstruction quality within the training domain.

\end{enumerate}

\section{Related Work}
\label{sec:related}

\textbf{Neural Volumetric Reconstruction.}
These models directly regress scene-level geometry from posed RGB images, departing from classical depth-fusion pipelines. Atlas~\cite{atlas2020eccv} established this paradigm through end-to-end signed distance function (SDF) prediction with feature back-projection and 3D convolution, followed by works that improved scalability and multi-view reasoning, such as NeuralRecon~\cite{neuralrecon2021cvpr} with recurrent sequential SDF fusion and VoRTX~\cite{vortx20213dv}.
As multi-view fusion matured, later methods focused on preserving surface details and improving efficiency: SimpleRecon~\cite{simplerecon2022eccv} leveraged strong multi-view depth priors with lightweight fusion, FineRecon~\cite{finerecon2023iccv} introduced resolution-agnostic supervision and depth-guided refinement, and DG-Recon~\cite{dgrecon2023iccv} and CVRecon~\cite{cvrecon2023iccv} addressed feature redundancy and empty-space ambiguity via geometry- and cost-volume-guided fusion. More recent works further explore adaptive feature modeling, hierarchical alignment, sparsification, structured representations, and feedback-driven mechanisms to enhance reconstruction fidelity, compactness, and efficiency~\cite{zuo2023ral, detailrefine2025icra, georecon2025ral, airplanes2024cvpr}. Meanwhile, semantic- and prior-guided approaches incorporate higher-level reasoning or explicit geometric constraints into volumetric pipelines~\cite{panorecon2024cvpr, cdrnet2024tcsvt, gprecon2025tvcg}, while efficient extraction strategies such as AniGrad~\cite{anigrad2025cvpr} reduce surface reconstruction cost through adaptive sampling guided by local surface complexity.

\textbf{Geometry-Aware Foundation Models.}
More recently, VFMs pretrained on large-scale data have increasingly been applied to geometry-centric tasks such as monocular depth estimation, multi-view geometry, and monocular 3D reconstruction. In monocular depth estimation, DepthCues~\cite{depthcues2025cvpr} analyzes the emergence of depth perception in pretrained vision models, while representative models such as Metric3D~\cite{metric3d2023iccv, metric3dv22024tpami}, the MoGe series~\cite{moge2025cvpr, moge22025neurips}, and the Depth Anything series~\cite{dav12024cvpr, dav22024neurips, dav32026iclr} improve zero-shot metric depth generalization, affine- or metric-scale 3D point-map prediction, and cross-domain robustness through canonical camera modeling, large-scale training, and distillation. For multi-view geometry, CroCo~\cite{croco2022neurips, crocov22023iccv} learns transferable geometry-aware representations through cross-view masked modeling, DUSt3R~\cite{dust3r2024cvpr} directly regresses point maps from image pairs without explicit calibration, and MASt3R and its extensions~\cite{mast3r2024eccv, mast3rsfm20253dv, mast3rslam2025cvpr} further improve matching accuracy and integrate learned priors into SfM and SLAM systems. Recent feed-forward geometry engines further extend these ideas to scalable and globally consistent reconstruction: 
VGGT~\cite{vggt2025cvpr} and VGGT-SLAM~\cite{vggtslam2025neurips} jointly infer camera parameters, point maps, and global alignment, $\pi^3$~\cite{pi32026iclr} mitigates reference-view bias via a permutation-equivariant formulation, and MapAnything~\cite{mapanything20263dv} supports metric 3D reconstruction with flexible conditioning on images, intrinsics, poses, and depth.

\section{Methodology}
\label{sec:method}



\begin{figure*}[t]
    \centering
    \includegraphics[width=1.0\linewidth]{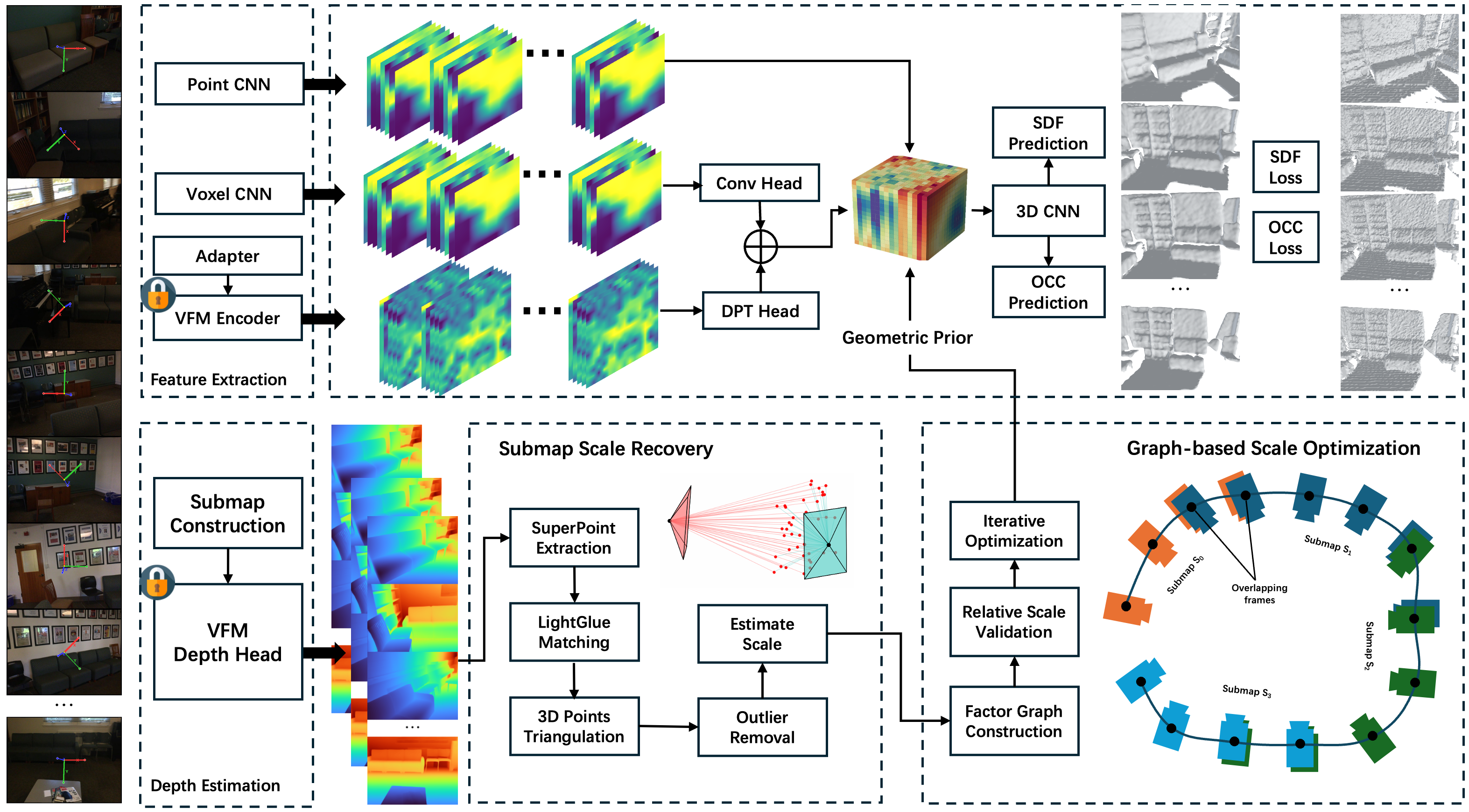}
    \caption{\textbf{System Overview.} 
    On the top is our \textit{VFM-augmented volumetric reconstruction network}, and in the bottom is our \textit{lightweight scale alignment module}.
    Our system divides the sequence into overlapping submaps and uses VFMs to predict scale-ambiguous depths. It then recovers per-submap scales via feature matching and triangulation, followed by factor-graph global scale optimization. The aligned depths are further refined with a fuse-and-back-projection process and are used as geometric priors for subsequent VFM-augmented neural reconstruction.}
    \label{fig:overview}
    \vspace{-2ex}
\end{figure*}

Given posed RGB images $\mathcal{I}=\{I_t\}_{t=1}^N$, camera poses $\mathcal{T}=\{\mathbf{T}_t\in SE(3)\}_{t=1}^N$, and intrinsics $\mathbf{K}$, our goal is to reconstruct a scene-level SDF volume $\mathcal{V}$ and extract a mesh $\mathcal{M}$ from its zero level set. We first select keyframes following~\cite{duzceker2021deepvideomvs}; all subsequent scale alignment, depth reprojection, and neural reconstruction are performed on these keyframes. \Cref{fig:overview} summarizes the full pipeline.

\subsection{Aligning Foundation Model Priors}
\label{sec::align}
We leverage the depth maps predicted by VFMs as dense geometric priors from the keyframes.
Within a single inference batch, the predicted depths are typically coherent up to a shared \emph{relative} scale, yielding geometrically plausible structure.
However, these VFMs are not trained with strict metric supervision, so their depth predictions are only defined up to an unknown \emph{sequence-dependent} scale.
As a result, VFM depths cannot be directly fused with camera poses with different scales (e.g., from SLAM/SfM), motivating a scale alignment stage that restores multi-view consistency.



We divide the keyframes into $M$ overlapping submaps ${\mathcal{S}_m}$ each with $n$ frames and $o$ shared frames between neighbors. Compared with previous splitting methods that share only a few views~\cite{fdslam2022icra}, the larger overlap provides more reliable inter-submap scale constraints. Each submap is processed independently by the VFM, yielding depths with an unknown positive scale $s_m$.

To recover global scale for each submap without needing a prior global map, we employ feature matching and triangulation. For each submap $\mathcal{S}_m$, we find correspondences between overlapping frames using SuperPoint~\cite{detone2018superpoint} and LightGlue~\cite{lindenberger2023lightglue}. Given camera poses $\mathbf{T}_t$ and feature correspondences, we triangulate 3D points and verify each triangulation through reprojection error checks and Chirality tests. Only points satisfying both constraints, i.e., positive depth along the optical axis and reprojection error below a threshold, are retained. 
\textcolor{black}{These filtered points anchor submap-wide scale without requiring matches in every frame, although entirely textureless submaps remain challenging, and provide depth values $z_t^p$ at pixels $u_t^p$.}
For each submap, we compute an initial scale factor by taking the robust median ratio between triangulated depths and the network predictions:

\begin{equation}
s_m^{(0)} = \mathop{\text{median}}_{p} \left\{ \frac{z_t^p}{\hat{D}_t^m(\mathbf{u}_t^p)} \right\}.
\end{equation}

\subsubsection{Graph-Based Scale Optimization}
The initial per-submap scales may still be noisy due to sparse triangulation and outliers. We therefore refine them by enforcing scale consistency between overlapping submaps. For each neighboring pair $(i,j)$, we compute a robust relative scale $r_{ij}$ as the median ratio between their VFM depth predictions on shared keyframes.

Since scale factors are inherently positive and multiplicative in nature, direct optimization in linear space risks negative or zero scales and poorly handles the multiplicative relationships between them. We therefore transform them to log-space where scale factors become additive. Let $x_i=\ln s_i$ and $\rho_{ij}=\ln r_{ij}$. We estimate globally consistent log-scales by solving
\begin{equation}
\begin{aligned}
\mathbf{x}^* =
\mathop{\arg\min}_{\mathbf{x}} \;
& \sum_{(i,j)\in\mathcal{E}} 
w_{ij}
\left(x_i - x_j - \rho_{ij}\right)^2 \\
& + \lambda \sum_i
\left(x_i - \ln s_i^{(0)}\right)^2,
\end{aligned}
\end{equation}
where $\mathcal{E}$ denotes overlapping submap pairs, $w_{ij}$ is the overlap confidence, and $\lambda$ regularizes the solution toward the triangulation-based initial scales. The final submap scales are $s_i^*=\exp(x_i^*)$.

\subsection{Depth Reprojection}
\label{sec::depth_reproj}

After obtaining the refined scale factors, we compute the globally-aligned depth maps for all keyframes:

\begin{equation}
D_{k_t}^{\text{aligned}}(\mathbf{u}) = s_m^* \cdot \hat{D}_{k_t}^m(\mathbf{u}), \quad \forall I_{k_t} \in \mathcal{S}_m, \forall \mathbf{u} \in \Omega.
\end{equation}

We then fuse these aligned depth maps into an initial SDF volume $\mathcal{V}_{\text{initial}}$ using volumetric fusion with running average. The fusion aggregates multi-view depth observations into a coherent 3D scene representation, reducing per-frame noise while preserving geometric details. For each keyframe $I_{k_t}$, we render a depth map $D_{k_t}^{\text{refined}}$ by first extracting mesh $\mathcal{M}_{\text{initial}}$ from the volume using marching cubes, then rendering new depth maps from the mesh using an off-the-shelf renderer:

\begin{equation}
D_{k_t}^{\text{refined}}(\mathbf{u}) = \text{render}\left(\mathcal{M}_{\text{initial}}, \mathbf{K}, \mathbf{T}_{k_t}, \mathbf{u}\right).
\end{equation}

\textcolor{black}{This projection enforces consistency with the fused geometry and can reduce noise and outliers, improving multi-view coherence, particularly in outdoor scenes.}
These refined depth maps, together with their corresponding keyframes $\mathcal{I}'$, serve as inputs to our VFM-augmented neural reconstruction pipeline.

\subsection{VFM-augmented Neural Reconstruction}
\label{sec:vfm_recon}

\begin{figure}[t]
    \centering
    \includegraphics[width=\linewidth]{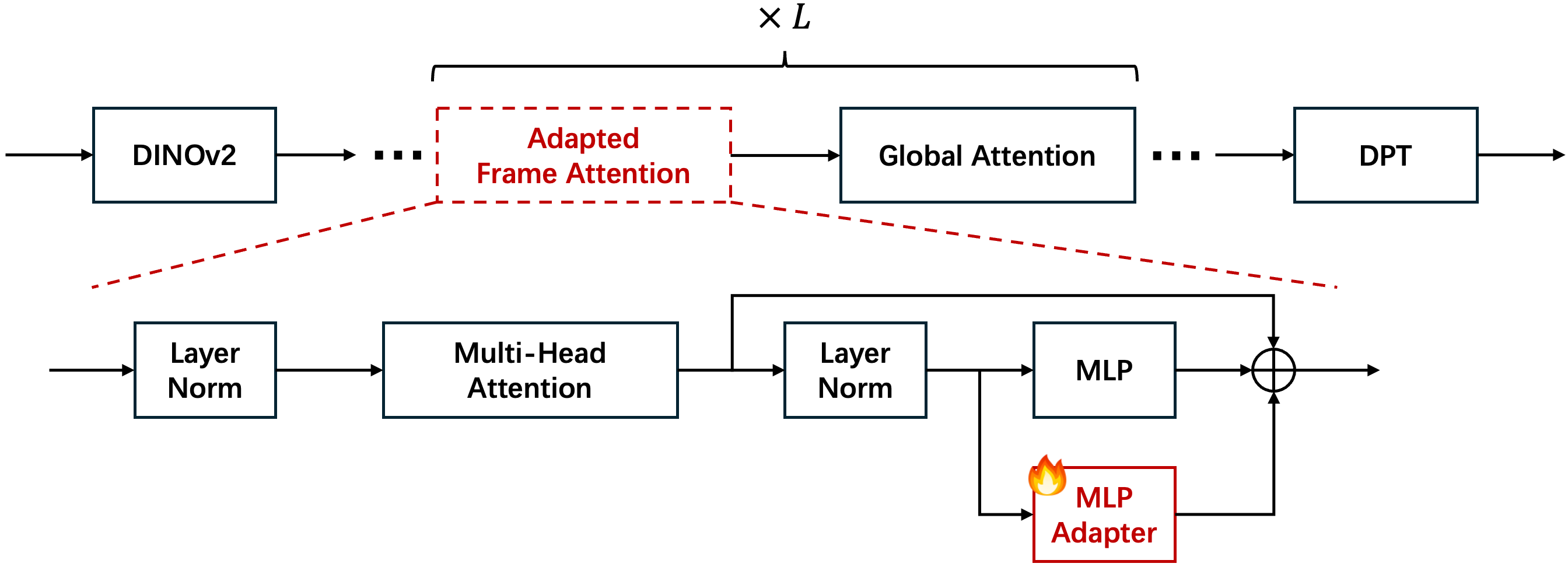}
    \caption{\textbf{Network Architecture.} Details of our adapted VFM-backbone for VFM-augmented volumetric reconstruction network, in which only the MLP adapter is trainable.}
    \vspace{-2ex}
    \label{fig:adapted-vggt}
\end{figure}


The scale-aligned and reprojected depths provide geometrically consistent priors, but they are still insufficient to directly produce high-quality implicit surfaces. We therefore use them as inputs to a neural volumetric reconstruction network, following the general refinement paradigm of FineRecon~\cite{finerecon2023iccv}. Unlike standard CNN-only pipelines, our network augments image features with task-adapted VFM features, which provide transferable geometry-aware representations for improved cross-domain reconstruction.

\subsubsection{Feature Extraction and Fusion}
\textcolor{black}{Given $P$ keyframes $\mathcal{I}'$ and refined depths $\mathcal{D}$, we extract geometry-aware VFM features by aggregating intermediate encoder representations through the DPT head. For CNN features, we directly adopt FineRecon's VoxelCNN and Point CNN, independent FPN-style EfficientNetV2-S encoders for voxel and query-point back-projection, respectively.}


\textcolor{black}{To adapt VFM features while retaining pretrained representations, we use a SelaVPR-inspired bottleneck MLP adapter~\cite{lu2024selavpr} with the backbone frozen. As shown in Fig.~3, the adapter runs parallel to the feed-forward MLP in frame-attention blocks, providing nonlinear residual updates without changing feature dimensions. This adapts within-frame representations while retaining pretrained cross-frame interaction modules. Empirically, adapting only frame-attention blocks outperforms adapting both frame- and global-attention blocks; we therefore use this configuration throughout.}

\textcolor{black}{The VFM features are channel-projected, bilinearly resized to the VoxelCNN resolution, and fused as $\mathcal{F}^{\mathrm{comb}}=\operatorname{LN}(\mathcal{F}^{\mathrm{voxel}})+\beta\operatorname{LN}(\widetilde{\mathcal{F}}^{\mathrm{vfm}})$, where 
$\mathcal{F}^{\mathrm{voxel}}$ and $\widetilde{\mathcal{F}}^{\mathrm{vfm}}$ denote CNN and aligned VFM features, respectively, and
$\beta$ is learnable. We then back-project the aggregated 2D features into a 3D volume using known camera parameters, averaging across views to construct the enhanced feature volume:}

\begin{equation}
\mathcal{F}_{\text{vol}}(\mathbf{x}_v) = \frac{1}{\sum_{t=1}^P w_{t,v}} \sum_{t=1}^P w_{t,v} \, \mathcal{F}_{k_t}^{\text{comb}}(\mathbf{u}_{t,v}),
\end{equation}
where $\mathcal{F}^{\text{comb}}$ is the fused VFM and CNN features, $w_{t,v} = \mathbf{1}_{[\mathbf{x}_v \in \text{frustum}_{k_t}]}$ indicates voxel visibility, and $\mathbf{u}_{t,v} = \pi(\mathbf{K}\mathbf{T}_{k_t}^{-1}\tilde{\mathbf{x}}_v)$ denotes the projected image coordinates with bilinear interpolation. This augmented feature volume, concatenated with the initial SDF reconstruction from the previous section, is processed by a 3D U-Net decoder:
\begin{equation}
\mathcal{V}_{\text{refined}} = F_\theta^{\text{dec}}\left(\left[\mathcal{F}_{\text{vol}}, \mathcal{V}_{\text{initial}}\right]\right),
\end{equation}
where $[\cdot, \cdot]$ denotes channel-wise concatenation and $F_\theta^{\text{dec}}$ represents a multi-level 3D convolutional decoder. 

In addition to the volumetric features, we also incorporate point back-projection features from~\cite{finerecon2023iccv}. For every 3D query point, we back-project the image features from the corresponding 2D location and aggregate multi-view features through average pooling. These point back-projection features are concatenated with $\mathcal{V}_{\text{refined}}$ to form a hybrid representation that combines both volumetric and point-wise information. The concatenated features are then processed through two separate MLPs to predict the signed distance field and occupancy probability, respectively.

\subsubsection{Neural Reconstruction}
The network is trained with ground-truth SDF fragments prepared following~\cite{finerecon2023iccv}. We use the same reconstruction objective, combining an occupancy BCE loss and an SDF regression loss:
\begin{equation}
\mathcal{L}_{\mathrm{total}}=\mathcal{L}_{\mathrm{occ}}+\mathcal{L}_{\mathrm{sdf}} .
\end{equation}
$\mathcal{L}_{\mathrm{occ}}$ supervises free/occupied space, while $\mathcal{L}_{\mathrm{sdf}}$ applies an $\ell_1$ loss on transformed SDF values at valid surface voxels.

\section{Experiments}
\label{sec:exp}

\subsection{Datasets, Baselines, and Metrics}
\label{sec::datasets}

\subsubsection{Datasets.}
We follow the standard protocol in neural volumetric reconstruction~\cite{neuralrecon2021cvpr, finerecon2023iccv, zuo2023ral, georecon2025ral} and use ScanNet~\cite{scannet2017cvpr} as our primary training and evaluation dataset. ScanNet is a large-scale indoor RGB-D dataset comprising 1,613 room-scale scans across 807 indoor scenes. We train and test our VFM-Recon using the official train/test splits.

To assess cross-domain generalizability, we follow Zuo \emph{et al.}~\cite{zuo2023ral} and evaluate on two additional out-of-distribution datasets: TUM RGBD~\cite{tumrgbd2012iros},  and Tanks and Temples~\cite{tat2017tog}. 
TUM RGBD consists of more constrained sequences
and exhibits distinct motion and viewpoint statistics, often featuring inward-facing trajectories with close-range observations; 
Tanks and Temples includes large-scale scenes spanning not only indoor spaces but also outdoor environments, with substantially higher appearance, geometry, and scale variability.

\subsubsection{Baselines.}
For in-domain evaluation, 
we compare with a broad range of neural reconstruction models, from earlier methods such as VoRTX~\cite{vortx20213dv}, NeuralRecon~\cite{neuralrecon2021cvpr}, SimpleRecon~\cite{simplerecon2022eccv}, FineRecon~\cite{finerecon2023iccv}, DG-Recon~\cite{dgrecon2023iccv}, and Zuo \textit{et al.}~\cite{zuo2023ral}, to recent SoTA systems including CDRNet\cite{cdrnet2024tcsvt}, GP-Recon~\cite{gprecon2025tvcg}, and GeoRecon~\cite{georecon2025ral}. To further demonstrate the reconstruction capability of the proposed model, we additionally report in-domain results from three recent neural reconstruction models, AirPlanes~\cite{airplanes2024cvpr}, PanoRecon~\cite{panorecon2024cvpr}, and DetailRefine~\cite{detailrefine2025icra}, as well as three geometry-aware VFMs, VGGT~\cite{vggt2025cvpr}, MapAnything~\cite{mapanything20263dv}, and $\pi^3$~\cite{pi32026iclr}.

For cross-domain evaluation, we compare with representative neural volumetric reconstruction methods, including NeuralRecon~\cite{neuralrecon2021cvpr}, VoRTX~\cite{vortx20213dv}, SimpleRecon~\cite{simplerecon2022eccv}, FineRecon~\cite{finerecon2023iccv}, Zuo \textit{et al.}\cite{zuo2023ral}, CDRNet~\cite{cdrnet2024tcsvt}, and GP-Recon~\cite{gprecon2025tvcg}, prioritizing reported results when available and otherwise running the official implementations under the same input setting. Moreover, we further evaluate four geometry-aware VFMs, including VGGT~\cite{vggt2025cvpr}, MoGe-2~\cite{moge22025neurips}, MapAnything~\cite{mapanything20263dv}, and $\pi^3$~\cite{pi32026iclr}, using their official implementations and pretrained weights to contextualize our performance and highlight the benefits of reconciling foundation priors with scale-consistent volumetric reconstruction.


\subsubsection{Metrics.} 
We follow the standard evaluation protocol in neural reconstruction~\cite{neuralrecon2021cvpr, simplerecon2022eccv, finerecon2023iccv, gprecon2025tvcg}. 
For reconstructed meshes, we report 3D metrics including Accuracy (\textit{Acc.}), Completeness (\textit{Comp.}), Chamfer distance (\textit{Cham.}), Precision (\textit{Prec.}), \textit{Recall}, and F1 score (\textit{F1}). 
For rendered depth, we additionally report 2D metrics: Absolute Relative error (\textit{Abs.Rel.}), Absolute Difference (\textit{Abs.Diff.}), Squared Relative error (\textit{Sq.Rel.}), Completion (\textit{Comp.}), and threshold accuracies (\textit{$\delta < 1.05$}, $\delta < 1.25$), where the $\delta$ metrics denote the percentage of pixels whose predicted depth is within a multiplicative threshold of the ground truth.

\subsection{Implementation Details}
\label{sec::imp_details}
We present two variants of the proposed model, 
\textcolor{black}{with VGGT~\cite{vggt2025cvpr} and $\pi^3$~\cite{pi32026iclr} as the VFM backbone, while the framework is also applicable to other geometry-aware VFMs.}
Both variants are equipped with a bottleneck MLP adapter with dimensions $1024 \rightarrow 512 \rightarrow 1024$. During finetuning, we freeze all pre-trained parameters of the VFM backbones and only optimize the adapter weights along with downstream task-specific heads. For VGGT, adapters are injected into all 24 frame-level attention blocks of its alternating aggregator, 
\textcolor{black}{operating in parallel with the feed-forward MLP through a scaled residual ($\alpha=0.2$).}
For $\pi^3$, adapters are inserted into 18 of its 36 decoder blocks (self-attention layers), while cross-attention layers remain frozen.

For scale alignment, 
\textcolor{black}{we empirically selected $n=16,o=4$ on ScanNet and $n=8,o=4$ on TUM RGB-D and Tanks and Temples, which yielded the best reconstruction performance among the configurations tested on each dataset.}
For depth reprojection, the globally aligned depth maps are fused via running-average volumetric fusion at a voxel resolution of $0.04$m. 
The point CNN and voxel CNN share the same FPN-style design~\cite{fpn2017cvpr} with an EfficientNetV2-S backbone~\cite{efficientnetv22021pmlr} but are trained independently, while the 3D CNN follows a U-Net architecture. We optimize the model with Adam using batch size 2, an initial learning rate of $1 \times 10^{-3}$, and a decay factor of $0.1$. We employ a cosine annealing schedule with the decay applied at the $80\%$ training milestone.

Training is conducted on two NVIDIA A6000 GPUs with 48\,GB memory each, and training takes approximately one week.
\textcolor{black}{At inference, the complete VGGT-based pipeline processes 100 ScanNet test scenes in approximately 50 minutes on one NVIDIA A6000, averaging 22 FPS with 24\,GB peak GPU memory.}

\subsection{Results and Discussion}
\label{sec::results}

\begin{table*}[t!]
  \caption{\textbf{2D Rendered Depth Results on ScanNet~\cite{scannet2017cvpr}.} The best, second best, and third best results are marked in \textbf{bold}, \underline{underlined}, and \textit{italic} respectively.}
  \vspace{-2ex}
  \label{tab:scannet2d}
  \centering
  \begin{tabular}{@{}lccccccc@{}}
    \toprule
    Method & Venue & Abs.Rel.$\downarrow$ & Abs.Diff.$\downarrow$ & Sq.Rel.$\downarrow$ & $\delta<{1.05}$$\uparrow$ & $\delta<{1.25}$$\uparrow$ & Comp.$\uparrow$ \\
    \midrule
    VoRTX~\cite{vortx20213dv} & 3DV'21 & 6.1 & 9.4 & 3.7 & {79.0} & 95.0 & 96.1 \\
    NeuralRecon~\cite{neuralrecon2021cvpr} & CVPR'21 & 6.5 & 10.6 & 3.1 & - & 94.8 & 90.9 \\
    SimpleRecon~\cite{simplerecon2022eccv} & ECCV'22 & \textbf{4.3} & 8.9 & \textbf{1.3} & 70.2 & \textbf{98.1} & - \\
    FineRecon~\cite{finerecon2023iccv} & ICCV'23 & 5.5 & 8.5 & 3.8 & \textit{82.8} & {95.8} & \underline{97.2} \\ 
    DG-Recon~\cite{dgrecon2023iccv} & ICCV'23 & 6.2 & 9.9 & 4.0 & - & 93.7 & \textit{96.6} \\
    Zuo \textit{et al.}~\cite{zuo2023ral} & RA-L'23 & 5.2 & {8.7} & \textit{2.5} & - & 94.8 & 90.6 \\
    CDRNet~\cite{cdrnet2024tcsvt} & TCSVT'24 & 6.2 & 8.4 & 3.7 & 76.4 & 94.8 & 96.1 \\
    GP-Recon~\cite{gprecon2025tvcg} & TVCG'25 & \underline{4.5} & \textbf{7.7} & \underline{2.2} & - & {95.4} & - \\
    GeoRecon~\cite{georecon2025ral} & RA-L'25 & 5.6 & 9.0 & 3.1 & - & 94.0 & 92.9 \\
    \midrule
    \textbf{VFM-Recon (VGGT)} & - & \textit{5.0} & \underline{8.1} & 3.7 & \textbf{85.5} & \underline{96.8} & \textbf{97.5}  \\
    \textbf{VFM-Recon ($\boldsymbol{\pi^3}$)} & - & \textit{5.0} & \textit{8.3} & 3.6 & \underline{83.4} & \textit{95.9} & \underline{97.2}  \\
    \bottomrule
  \end{tabular}
  \vspace{-3ex}
\end{table*}

\begin{table}[t]
  \caption{\textbf{3D Reconstruction Results on ScanNet~\cite{scannet2017cvpr}.} The best, second best, and third best results are marked in \textbf{bold}, \underline{underlined}, and \textit{italic} respectively. 
  }
  \vspace{-2ex}
  \label{tab:scannet3d}
  \centering
  \resizebox{\linewidth}{!}{
  \begin{tabular}{@{}lcccccc@{}}
    \toprule
    Method 
    & Acc.$\downarrow$ & Comp.$\downarrow$ & Cham.$\downarrow$ & Prec.$\uparrow$ & Recall$\uparrow$ & F1$\uparrow$ \\
    \midrule
    VoRTX~\cite{vortx20213dv} 
    & 5.4 & 9.0 & - & 70.8 & 58.8 & 64.1 \\
    NeuralRecon~\cite{neuralrecon2021cvpr}
    & 5.4 & 12.8 & - & 68.4 & 47.9 & 56.2 \\
    SimpleRecon~\cite{simplerecon2022eccv}
    & 6.1 & {5.5} & 5.8 & 68.6 & {65.8} & 67.1 \\
    FineRecon~\cite{finerecon2023iccv} 
    & 5.6 & \textit{5.4} & \textit{5.5} & \underline{76.9} & \textit{71.3} & \underline{73.6} \\ 
    DG-Recon~\cite{dgrecon2023iccv} 
    & \textbf{3.9} & 6.8 & \underline{5.4} & \underline{76.9} & 63.6 & {69.4} \\
    Zuo \textit{et al.}~\cite{zuo2023ral} 
    & 5.8 & 11.0 & - & 66.5 & 50.5 & 57.2 \\
    AirPlanes~\cite{airplanes2024cvpr} 
    & \textit{4.9} & 6.0 & \underline{5.4} & 70.8 & 62.2 & 66.2 \\ 
    PanoRecon~\cite{panorecon2024cvpr} 
    & 6.4 & 8.9 & - & 65.6 & 53.0 & 58.4 \\
    CDRNet~\cite{cdrnet2024tcsvt} 
    & 6.8 & 6.2 & - & 60.9 & 61.6 & 61.2 \\
    GeoRecon~\cite{georecon2025ral} 
    & 5.6 & 9.8 & 7.7 & 68.3 & 53.0 & 59.5 \\
    GP-Recon~\cite{gprecon2025tvcg} 
    & \textbf{3.9} & 9.3 & 6.6 & \textbf{77.6} & 61.0 & 68.1 \\
    DetailRefine~\cite{detailrefine2025icra} 
    & \underline{4.4} & 9.3 & 6.9 & 70.9 & 54.1 & 61.3 \\
    \midrule
    VGGT~\cite{vggt2025cvpr} & - & - & - & 48.4 & 40.3 & 43.4 \\
    $\pi^3$~\cite{pi32026iclr} & - & - & - & \textit{54.9} & \textit{53.9} & \textit{54.0} \\
    MapAnything~\cite{mapanything20263dv} & - & - & - & 41.9 & 27.5 & 31.4 \\
    \midrule
    \textbf{VFM-Recon (VGGT)} 
    & 5.6 & \textbf{4.7} & \textbf{5.1} & \textit{75.7} & \textbf{73.2} & \textbf{74.3} \\
    \textbf{VFM-Recon ($\boldsymbol{\pi^3}$)} 
    & 5.6 & \underline{5.1} & 5.6 & {75.2} & \underline{71.7} & \textit{73.5} \\
  \bottomrule
  \end{tabular}
  }
  \vspace{-3ex}
\end{table}

\begin{table}[t]
  \caption{
  \textbf{Cross-Domain Evaluation of Reconstructed Mesh.} The best, second best, and third best results are marked in \textbf{bold}, \underline{underlined}, and \textit{italic} respectively. 
  }
  \vspace{-2ex}
  \label{tab:crossdomain}
  \centering
  \resizebox{\linewidth}{!}{
  \begin{tabular}{@{}lcccccc@{}}
    \toprule
    \multirow{2}{*}{Method} 
    & \multicolumn{3}{c}{TUM RGBD~\cite{tumrgbd2012iros}} & \multicolumn{3}{c}{Tanks\&Temples~\cite{tat2017tog}} \\ 
    & Prec.$\uparrow$ & Recall$\uparrow$ & F1$\uparrow$ & Prec.$\uparrow$ & Recall$\uparrow$ & F1$\uparrow$ \\
    \midrule
    VoRTX~\cite{vortx20213dv} 
    & 28.0 & 19.4 & 22.9 & - & - & - \\ 
    NeuralRecon~\cite{neuralrecon2021cvpr}
    & \underline{49.4} & 24.8 & 32.4 & 42.6 & 24.7 & 30.2 \\
    SimpleRecon~\cite{simplerecon2022eccv} 
    & 40.0 & \underline{48.0} & \textit{43.5} & 44.4 & \textit{31.5} & {35.7} \\
    Zuo \textit{et al.}~\cite{zuo2023ral} 
    & 36.9 & 13.3 & 19.5 & - & - & - \\ 
    FineRecon~\cite{finerecon2023iccv} 
    & 27.4 & 32.2 & 29.6 & {47.0} & 26.5 & 33.1 \\
    CDRNet~\cite{cdrnet2024tcsvt} 
    & 28.1 & 31.7 & 29.6 & 39.6 & 25.4 & 29.2 \\
    GP-Recon~\cite{gprecon2025tvcg} 
    & \textit{48.9} & {32.8} & {38.6} & - & - & - \\
    \midrule
    VGGT~\cite{vggt2025cvpr} 
    & 25.2 & 19.9 & 21.9 & {51.0} & {55.8} & {51.8} \\
    MoGe-2~\cite{moge22025neurips} 
    & 27.3 & \textit{33.5} & 29.9 & 41.7 & 39.2 & 40.3 \\
    MapAnything~\cite{mapanything20263dv} 
    & \textbf{59.2} & 32.5 & 41.3 & \underline{70.6} & 38.4 & 45.2 \\
    $\pi^3$~\cite{pi32026iclr} 
    & 34.4 & 31.2 & 29.8 & \textit{69.7} & \textit{67.0} & \textit{67.0}  \\
    \midrule
    \textbf{VFM-Recon (VGGT)} 
    & {43.6} & \textbf{50.7} & \textbf{46.7} & {68.6} & \underline{73.7} & \underline{70.1} \\
    \textbf{VFM-Recon ($\boldsymbol{\pi^3}$)} 
    & 42.6 & \underline{48.0} & \underline{45.6} & \textbf{73.2} & \textbf{74.0} & \textbf{73.5} \\
  \bottomrule
  \end{tabular}  
  }
  \vspace{-2ex}
\end{table}

\subsubsection{Quantitative Comparisons.}

The quantitative results are summarized in \Cref{tab:scannet2d,tab:scannet3d,tab:crossdomain}. 

On the in-domain ScanNet benchmark~\cite{scannet2017cvpr}, VFM-Recon achieves strong performance on both rendered depth and reconstructed mesh evaluation. 
For rendered depth, VFM-Recon (VGGT) obtains the best $\delta<1.05$ and \textit{Comp.} scores, while maintaining competitive \textit{Abs.Rel.} and \textit{Abs.Diff.}. 
For 3D reconstruction, both VFM-Recon variants show clear advantages on completeness-oriented and overall metrics, with VFM-Recon (VGGT) achieving the best \textit{Comp.}, \textit{Cham.}, \textit{Recall}, and \textit{F1} scores. 
Although the \textit{Acc.} and \textit{Sq.Rel.} results are relatively weaker, these metrics are sensitive to outliers, spurious fragments, and slight surface thickening, which may occur when improving surface coverage and recovering fine structures. 
Compared with directly using geometry-aware VFMs, VFM-Recon substantially improves reconstruction quality by adapting foundation priors into a scale-consistent volumetric framework; for example, the F1 score improves from 43.4 to 74.3 with VGGT~\cite{vggt2025cvpr} and from 54.0 to 73.5 with $\pi^3$~\cite{pi32026iclr} on ScanNet. 

Under cross-domain settings, VFM-Recon also generalizes well to TUM RGBD~\cite{tumrgbd2012iros} and Tanks and Temples~\cite{tat2017tog}. 
On TUM RGBD, VFM-Recon (VGGT) achieves the best \textit{Recall} and \textit{F1}, while on the more challenging Tanks and Temples benchmark, VFM-Recon ($\pi^3$) achieves the best overall performance with 73.2 \textit{Prec.}, 74.0 \textit{Recall}, and 73.5 \textit{F1}. 
These results show that directly applying pretrained VFMs is insufficient for high-quality mesh reconstruction, whereas integrating VFM priors with scale-consistent volumetric fusion leads to more complete and robust reconstruction across both in-domain and cross-domain scenarios.

\subsubsection{Qualitative Comparisons.}

\begin{figure*}[t]
    \centering
    \includegraphics[width=\linewidth]{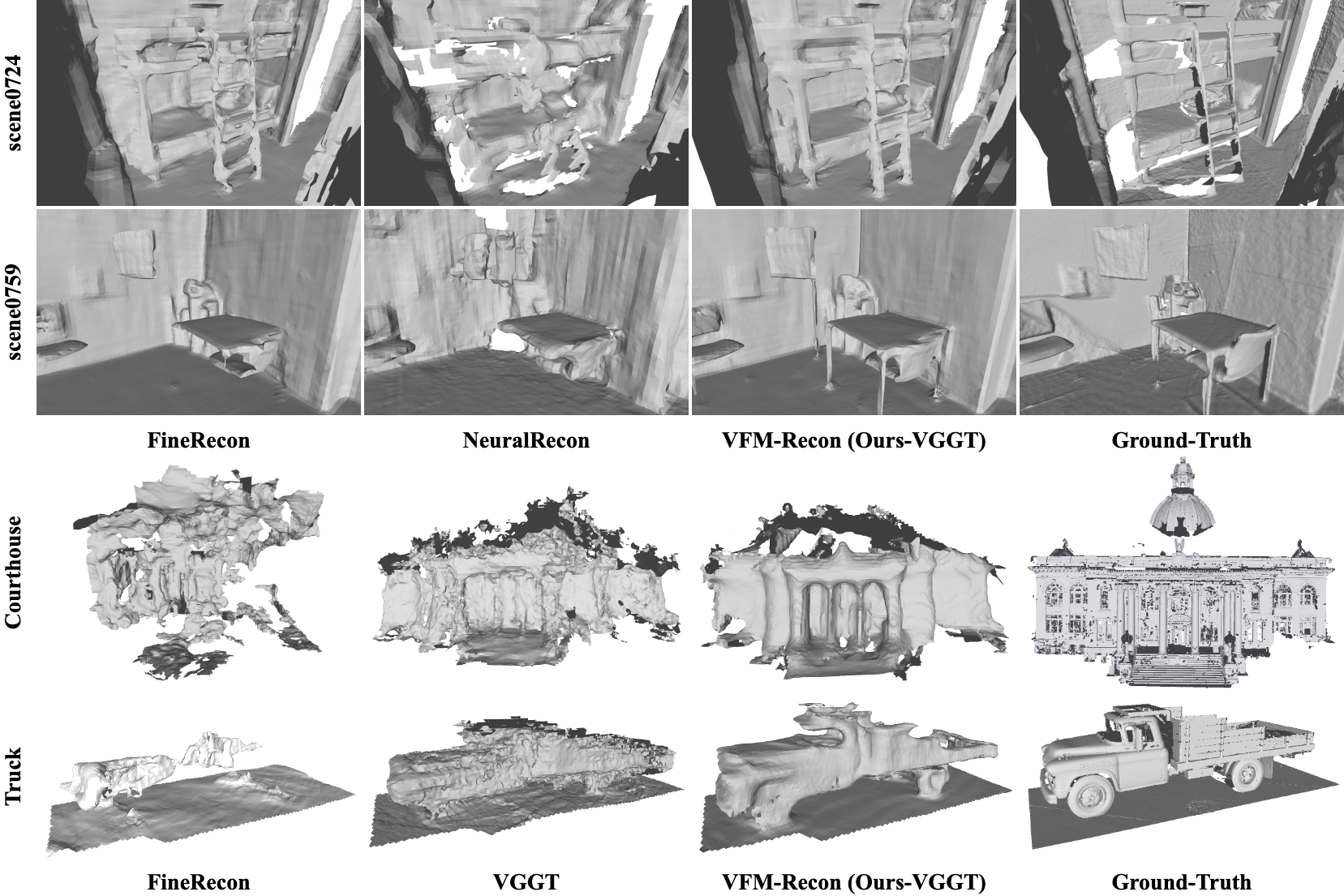}
    \caption{\textbf{Qualitative Comparisons.} In the in-domain examples (top 2 rows), VFM-Recon is compared with FineRecon~\cite{finerecon2023iccv} and NeuralRecon~\cite{neuralrecon2021cvpr}, showing improved structural details such as staircases, table legs, and TV surfaces. In the cross-domain examples (bottom 2 rows), VFM-Recon is compared with FineRecon~\cite{finerecon2023iccv} and VGGT~\cite{vggt2025cvpr}, producing more complete structures with fewer fragmented artifacts in challenging outdoor scenes.}
    \label{fig:qualitative}
\end{figure*}

The qualitative comparisons in \Cref{fig:qualitative} further support the quantitative results. 

In the in-domain ScanNet examples, VFM-Recon reconstructs finer and cleaner structures than NeuralRecon~\cite{neuralrecon2021cvpr} and FineRecon~\cite{finerecon2023iccv}, including clearer staircases, more complete table legs, and smoother TV or wall surfaces. 
These improvements are consistent with the gains on \textit{Comp.}, \textit{Cham.}, \textit{Recall}, and \textit{F1}, indicating that VFM-Recon improves surface coverage while preserving structural details. 

In the cross-domain Tanks and Temples~\cite{tat2017tog} examples, FineRecon suffers from severe incompleteness under domain shift, while VGGT~\cite{vggt2025cvpr} produces plausible but noisy and fragmented geometry. 
In contrast, VFM-Recon reconstructs more coherent and complete surfaces with fewer spurious artifacts, such as the facade structure in \textit{Courthouse} and the object silhouettes in \textit{Truck}. 
\textcolor{black}{Nevertheless, a gap to ground truth remains in challenging outdoor scenes. Although VFM-Recon improves structural continuity and completeness, persistent over-smoothing, missing details, and local distortions may reflect inaccurate depth predictions and limited spatial resolution. These limitations could be addressed through consistency-aware fusion that down-weights unreliable depth observations, together with coarse-to-fine reconstruction using finer voxels and high-resolution features where sufficient multi-view evidence is available.}

\label{sec::ablation}
\begin{table*}[t]
  \caption{
  \textbf{Ablation Studies.} V.D. denotes the use of VGGT depth, S.A. and D.R refer to our scale alignment and depth reprojection step respectively, A.V. indicates task-adapted VGGT features.
  The best, second best, and third best results are marked in \textbf{bold}, \underline{underlined}, and \textit{italic} respectively. }
  \vspace{-2ex}
  \label{tab:ablation}
  \centering
  \begin{tabular}{@{}l@{\hspace{1pt}}cccc@{\hspace{12pt}}ccc@{\hspace{12pt}}ccc@{}}
    \toprule
    \multirow{2}{*}{Variant} & \multicolumn{4}{c}{Settings} & \multicolumn{3}{c}{ScanNet~\cite{scannet2017cvpr}} & \multicolumn{3}{c}{Tanks\&Temples~\cite{tat2017tog}} \\ 
    & V.D. & S.A. & D.R. & A.V. & Prec.$\uparrow$ & Recall$\uparrow$ & F1$\uparrow$ & Prec.$\uparrow$ & Recall$\uparrow$ & F1$\uparrow$ \\
    \midrule
    Default & $\checkmark$ & $\checkmark$ & $\checkmark$ & $\checkmark$ & \underline{75.7} & \underline{73.2} & \underline{74.3} & \textbf{68.6} & \textbf{73.7} & \textbf{70.1} \\
    \midrule
    w/o S.A. & $\checkmark$ &  & $\checkmark$ & $\checkmark$ & 74.5 & 71.5 & 72.8 & 45.5 & \underline{48.4} & 45.4 \\
    w/o D.R. & $\checkmark$ & $\checkmark$ &  & $\checkmark$ & \textbf{76.5} & \textbf{75.4} & \textbf{75.8} & \underline{59.8} & 48.3 & \underline{50.5} \\
    w/o S.A.\&D.R. & $\checkmark$ &  &  & $\checkmark$ & 73.7 & 71.0 & 72.3 & 49.0 & 32.6 & 36.0 \\ 
    Adapted VGGT Only &  &  &  & $\checkmark$ & \underline{75.7} & 72.2 & 73.9 & 42.8 & 31.6 & 33.7 \\
    Naive VGGT Only &  &  &  &  & 73.5 & 70.8 & 72.1 & 42.5 & 31.7 & 33.4 \\
    \bottomrule
  \end{tabular}
\end{table*}

\subsection{Ablation Study}
To validate our design choices, we ablate the key components of VFM-Recon in \Cref{tab:ablation}. All ablation experiments are performed using the VGGT-based variant of VFM-Recon, so that the influence of different components can be analyzed without introducing additional backbone-related factors. 
Overall, removing any component degrades cross-domain performance (see \textit{w/o S.A.}, \textit{w/o S.A.\&D.R.}, \textit{Adapted VGGT Only}, and \textit{Naive VGGT Only}), confirming that the proposed modules contribute complementary benefits. 
Interestingly, solely removing depth reprojection (the \textit{w/o D.R.} row) slightly improves in-domain results on ScanNet~\cite{scannet2017cvpr} but causes a much larger degradation under domain shift. This indicates that depth reprojection provides stronger cross-domain regularization by enforcing multi-view consistency in a geometry-grounded manner. 
\textcolor{black}{Although this single-pass fusion--rendering stage can propagate initial geometric errors and introduce discretization artifacts, the ablation shows a modest in-domain cost and a substantial cross-domain benefit.}

\section{Conclusion}
\label{sec:conclusion}
We presented \textbf{VFM-Recon}, a novel framework that bridges between transferable VFM priors and the scale consistency required by neural volumetric reconstruction. Our key insight is to decouple cross-domain reconstruction into two complementary stages: (i) a lightweight geometric alignment module that resolves the scale ambiguity and inter-frame inconsistency of VFMs depth predictions, enabling stable multi-view fusion, and (ii) a VFM-augmented neural reconstruction pipeline that integrates task-adapted foundation features to improve robustness under domain shift. Extensive experiments on in-domain as well as cross-domain evaluations demonstrate that VFM-Recon achieves SoTA performance, with particularly strong gains on challenging cross-domain outdoor scenes. We believe VFM-Recon offers a practical and general recipe for leveraging foundation priors in scene-level reconstruction, opening up new opportunities for robust 3D perception across diverse environments.




\bibliographystyle{IEEEtran}
\bibliography{ref}

\vfill

\end{document}